\documentclass{article}
\usepackage{ijcai21}

\usepackage{times}
\usepackage{soul}
\usepackage{url}
\usepackage[hidelinks]{hyperref}
\usepackage[utf8]{inputenc}
\usepackage[small]{caption}
\usepackage{amsthm}
\usepackage{booktabs}
\usepackage{algorithm,algorithmic,amsmath,amsfonts}
\usepackage{graphicx,psfrag,epsfig,subfigure}
\usepackage{xcolor}
\usepackage{booktabs}
\usepackage{hyperref}
\hypersetup{
    colorlinks=true,
    linkcolor=red,
    citecolor=blue,
}
\usepackage{multirow}
\title{Reducing Adversarial Training Cost with Gradient Approximation}

\author{
    Huihui Gong
    \affiliations
    The University of Sydney
    \emails
    hgon9611@uni.sydney.edu.au
}

\begin{document}

\maketitle

\begin{abstract}
Deep learning models have achieved state-of-the-art performances in various domains, while they are vulnerable to the inputs with well-crafted but small perturbations, which are named after adversarial examples (AEs). Among many strategies to improve the model robustness against AEs, Projected Gradient Descent (PGD) based adversarial training is one of the most effective methods. Unfortunately, the prohibitive computational overhead of generating strong enough AEs, due to the maximization of the loss function, sometimes makes the regular PGD adversarial training impractical when using larger and more complicated models. In this paper, we propose that the adversarial loss can be approximated by the partial sum of Taylor series. Furthermore, we approximate the gradient of adversarial loss and propose a new and efficient adversarial training method, adversarial training with gradient approximation (GAAT), to reduce the cost of building up robust models. Additionally, extensive experiments demonstrate that this efficiency improvement can be achieved without any or with very little loss in accuracy on natural and adversarial examples, which show that our proposed method saves up to 60\% of the training time with comparable model test accuracy on MNIST, CIFAR-10 and CIFAR-100 datasets.
\end{abstract}

\section{Introduction}

In recent years, deep learning models have given a mightily impressive performance in a variety of domains including automatic driving, machine translation, automatic speech recognition, cyber defence, and so on. However, these deep networks are extremely vulnerable to particularly designed inputs with imperceptible perturbations, named \textit{adversarial examples}. In order to defend against the threat of adversarial examples, relevant researchers proposed numerous robust deep networks, which usually include main types, provable defences \cite{Cohen2019,Salmany2019,Singla2020} and empirical defences \cite{Kurakin2016,Madry2018}. Among the proposed methods, \textit{adversarial training}, a famous empirical defence, is one of the most effective methods \cite{Cai2018,Wang2019,Hendrycks2019}. Generally, in an adversarial training epoch, the adversarial versions of natural training data are generated and then they are used to train the model to increase its robustness towards such adversaries.

\begin{table}[t]
\centering
\begin{tabular}{lrr}
\toprule[1.5pt]
Model               & Std Training  & RAT     \\
\midrule
VGGNet-16           & 31 min        & 421 min \\
ResNet-18           & 17 min        & 189 min  \\
ResNet-50           & 41 min        & 563 min \\
WideResNet-28x4     & 94 min        & 1345 min \\
\bottomrule[1.5pt]
\end{tabular}
\caption{Time consumed by standard training and RAT of different models for the CIFAR-10 dataset. All experiments are run with 200 epochs and the same hyper-parameters. RAT is conducted using the following extra parameters: \{PGD steps, step size, maximum perturbation\}=\{10, 1/255, 8/255\}.\label{tab.std}}
\end{table}

The bulk of research interests about adversarial examples are relevant to generating stronger adversarial examples as well as improving adversarial robustness of deep models, while less attention is put into the cost of training a robust model. According to Table \ref{tab.std}, the regular adversarial training (RAT) is much more costly than the standard training when finishing the CIFAR-10 image classification task with four common deep models. Here, we use the same hardware, training dataset, training hyper-parameters and total number of epochs. Thus, it is contributive to reduce the computational cost of adversarial training. Nowadays, a robust method relates to training an ensemble of deep models \cite{Strauss2017,Tramer2018,Pang2019}, which greatly reduces the training time of each model. In \cite{Shafahi2019}, the authors simultaneously generated adversarial examples and updated model parameters to reduce the training epochs with comparable accuracy performances.

In this paper, we propose a novel adversarial training method, \textit{adversarial training with gradient approximation} (GAAT), which significantly reduces the cost of adversarial training by considering the gradient approximation of adversarial loss. Specifically, in consideration of the imperceptibility of the adversarial perturbation, we leverage the standard loss, its gradient, second-order derivative with respect to the inputs as well as the adversarial perturbation to approximate the adversarial loss. Then, we utilize it to generate strong adversarial examples. Finally, the generated adversarial examples are used for adversarial training to improve the robustness of deep models.

\textbf{Our contributions.} We propose a simple but effective method, adversarial training with gradient approximation (GAAT), which trains the robust deep models faster than RAT. The key idea of GAAT is to replace the costly gradient of adversarial loss with the approximate gradient. Based on popular deep learning models (shallow neural networks \cite{Wong2018}, VGGNet \cite{Simonyan2014}, ResNet \cite{He2015} and WideResNet \cite{Zagoruyko2016}), our proposed robust models trained on MNIST, CIFAR-10 and CIFAR-100 datasets save up to 60\% of training time with comparable robust performances to models trained with RAT. Moreover, we compare our method with the delayed adversarial training \cite{Gupta2020} and combine them to further improve the training efficiency.

\section{Background and Related Work}
Since proposed in 2014 \cite{Szegedy2014}, adversarial examples have become a hot interest among machine learning researchers and scholars. Until now, many papers have studied adversarial attacks and robust models. In this section, we only emphasize several methods and concepts that are most relevant to our work.

\textbf{Regular adversarial training.} Initially, the most commonly utilized method to create adversarial examples is based on projected gradient descent (PGD), which originally referred to the basic iteration method proposed by Kurakin {\it et al.}~\shortcite{Kurakin2017}. To improve the previous methods, multiple stronger threat models are proposed, e.g., Madry {\it et al.}~\shortcite{Madry2018} proposed the famous PGD approach, which is regarded as the most effective first-order attack. Interestingly, this method can transfer into a heuristic defence by exploiting the adversarial examples in adversarial training, which serves as an effective defence method in practice. Concretely speaking, let an image $x$, its label $y$ and some ball $\mathcal{B}(x,\epsilon)$ around $x$ with radius $\epsilon$. The PGD attack method is formulated as the following iteration:
\begin{equation}\label{eqn.pgd}
\delta^{t+1} = \Pi\Big(\delta^t+\alpha\cdot \mathrm{sign}\left(\nabla_{\delta^t}\ell(f_\theta(x+\delta^t),y)\right)\Big),
\end{equation}
where $\delta^0$ is the initial perturbation (usually set to zero or a random value), $\alpha$ is the step size and $\Pi(\cdot)$ projects the results of the gradient step into the $\epsilon$-ball. In this paper, we consider the $l_\infty$ ball $\mathcal{B}_\infty(x,\epsilon)=\{x+\delta: ||\delta||_\infty\le\epsilon\}$. $\mathrm{sign}(\cdot)$ is used to make the iteration converge faster, because the standard gradient steps $[\nabla_{\delta^t}\ell(f_\theta(x+\delta^t),y)]$ are typically too small. If $t+1=T$, it is denoted as a $T$-step PGD attack. The bigger the value of $T$ is, the stronger the adversarial examples are, resulting in a greater chance of misclassification by a well-trained classifier $f_\theta$. In addition to its effectiveness of threatening a trained classifier, PGD attack also enlightens us on how to defend against such strong first-order adversaries. Specifically, instead of minimizing the loss function evaluated at a natural example $x$, we minimize the loss function on an adversarial example ($x+\delta^T$), where $\delta^T$ is generated by the $T$-step PGD attack (\ref{eqn.pgd}) for some ball $\mathcal{B}(x,\epsilon)$. We refer to the above PGD based adversarial training as the regular adversarial training (RAT) and summarize it in Algorithm \ref{alg.pgd}.

\begin{algorithm}[t]
\caption{Regular adversarial training (RAT)}
\label{alg.pgd}
\textbf{Input}:  $M$ batches of natural images $(x,y)$\\
\textbf{Parameter}: Training epochs $N$, trained model $f_{\theta}$, learning rate $\gamma$, PGD parameters \{$T$,$\alpha$,$\epsilon$\}\\
\textbf{Output}: Robust model parameters $\theta$\\
\begin{algorithmic}[1]
\FOR{$n=1\dots N$}
\FOR{$i=1\dots M$}
\STATE {\small\color{gray} \textit{\# Conduct PGD adversarial attack}}
\STATE {$\delta=0$}  {\small\color{gray} \textit{\# or random value}}
\FOR{$t=1\dots T$}
\STATE {$\delta=\delta+\alpha\cdot\mathrm{sign}(\nabla_\delta\ell(f_\theta(x_i+\delta),y_i))$}
\STATE {$\delta=\max(\min(\delta,\epsilon),-\epsilon)$}  {\small\color{gray} \textit{\# projected into $l_\infty$ ball}}
\ENDFOR
\STATE {\small\color{gray} \textit{\# Update classifier parameters with some optimizer}}
\STATE {$\theta=\theta-\gamma\nabla_\theta\ell(f_\theta(x_i+\delta),y_i)$}
\ENDFOR
\ENDFOR
\RETURN {$\theta$} {\small\color{gray} \textit{\# return parameters of the robust classifier}}
\end{algorithmic}
\end{algorithm}

\begin{figure*}[t]
\centering
\psfrag{R(x,del)}[c][c][0.7]{$R(x,\delta)$}
\psfrag{epoch}[c][c][0.7]{Epoch}
\psfrag{1 Step}[c][c][0.6]{1 Step}
\psfrag{4 Steps}[c][c][0.6]{4 Steps}
\psfrag{10 Steps}[c][c][0.6]{10 Steps}
\psfrag{16 Steps}[c][c][0.6]{16 Steps}
\subfigure[ResNet-18]{\includegraphics[width=0.82\columnwidth]{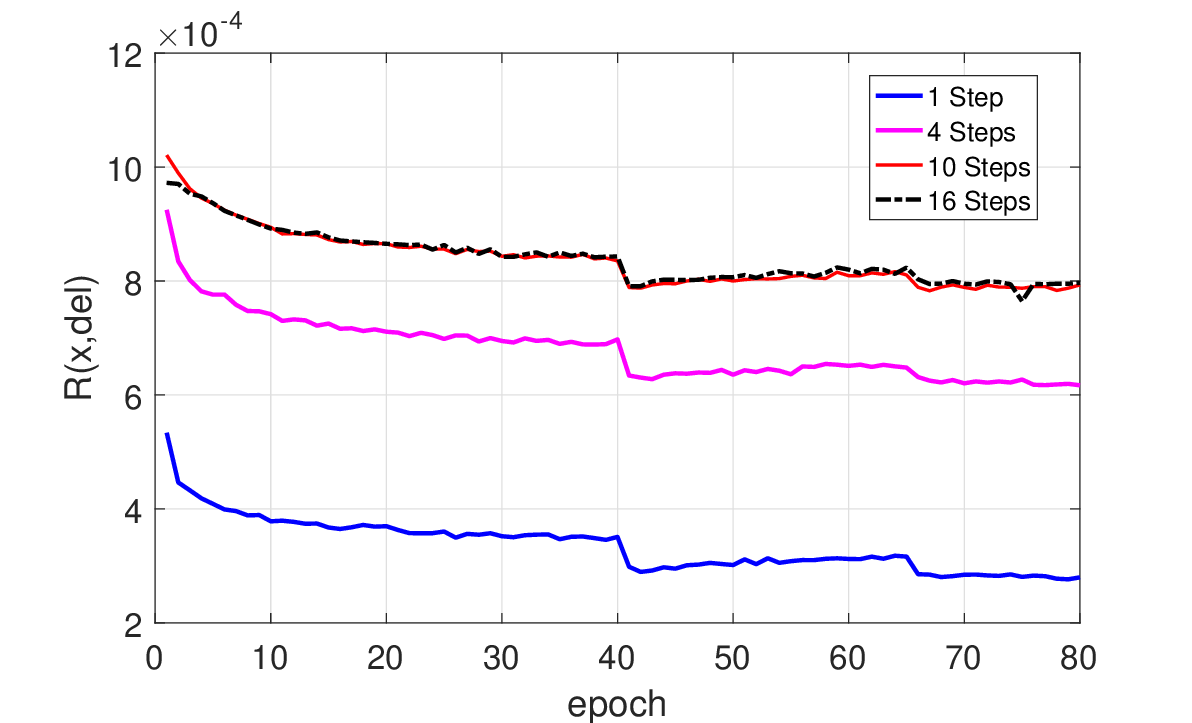}}
\subfigure[ResNet-50]{\includegraphics[width=0.82\columnwidth]{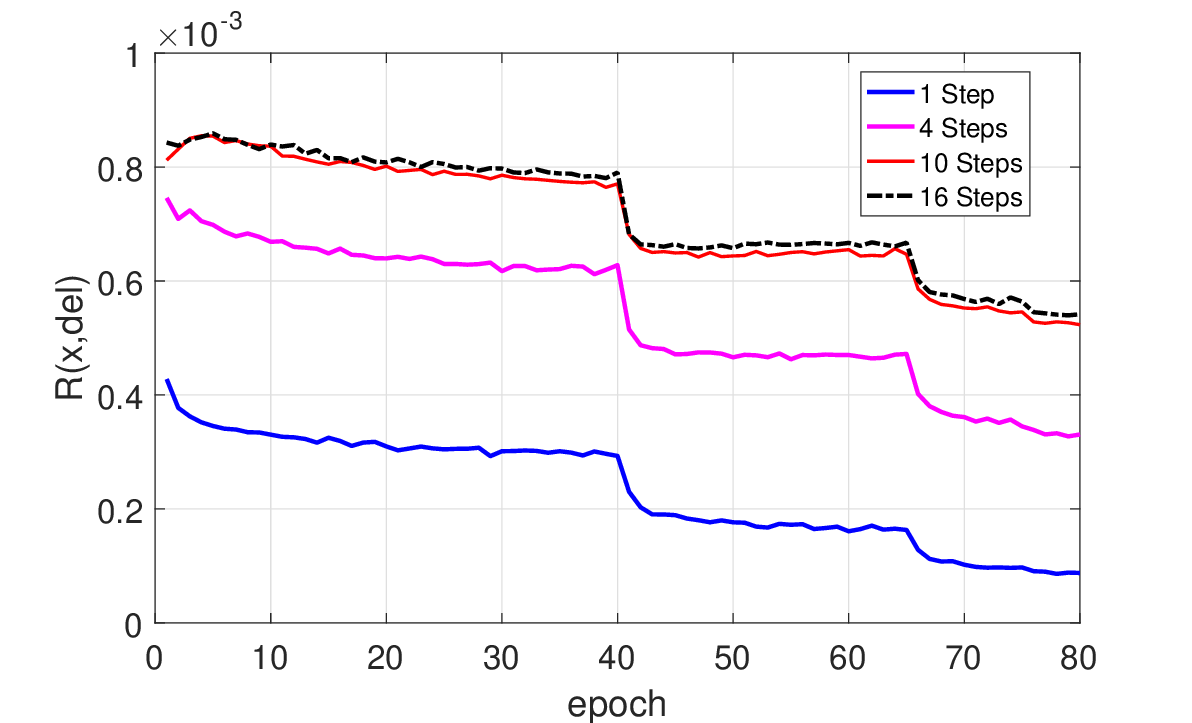}}
\caption{The value of $R(x,\delta)$ of different steps PGD regular adversarial training for two ResNet models  on the CIFAR-10 dataset.}
\label{fig.obser}
\end{figure*}

\textbf{Improving efficiency of adversarial training.} Another relevant work is about reducing the overhead of adversarial training.  Shafahi {\it et al.}~\shortcite{Shafahi2019} took advantage of gradient information from model updates to reduce the cost of generating adversaries and claimed about 7 times faster than RAT. However, one is required to run their methods for multiple times to select better models that reduces the overall training time, which is significant because the suboptimal models achieve the suboptimal  generalization and robustness \cite{Gupta2020}. Wong {\it et al.}~\shortcite{Wong2020} proposed the a fast adversarial training (Fast AT) method that reuses the well-known method, fast gradient sign method \cite{Goodfellow2015}. By initializing the perturbation with a random value, Fast AT is as effective as RAT but has much lower cost. Nevertheless, Fast AT has its limitation, i.e., it sometimes fails due to catastrophic overfitting, as illustrated in \cite{Wong2020,Li2020}. Gupta {\it et al.}~\shortcite{Gupta2020} showed that in adversarial training, using natural examples for the initial epochs and adversarial samples for the rest epochs achieve comparable robustness with RAT, but this method, denoted as delayed adversarial training (DAT), consumes much less time and computational resources.

\textbf{Taylor series.} Last but not least, our paper relies on Taylor series, which is an infinite sum of terms that are calculated from derivatives of a function at a single point. Note that the notation $C^m[a,b]$ denotes the set of all functions $f(\cdot)$ so that $f(\cdot)$ and its first, second, ..., $n$th-order derivatives are continuous at the computational interval $[a,b]$, the Taylor series of a function $f(x)$ is given by
\begin{equation*}
\begin{aligned}
f(x+\triangle x)=&f(x)+\frac{1}{1!}\triangle xf^{(1)}(x)+\frac{1}{2!}\triangle x^2f^{(2)}(x)\\
&+\frac{1}{3!}\triangle x^3f^{(3)}(x)+\cdots+\frac{1}{n!}\triangle x^nf^{(n)}(x),
\end{aligned}
\end{equation*}
where $f^{(i)}$ denotes the $i$th-order derivative of $f(x)$. In mathematics and engineering fields, like future minimization \cite{Zhang2017} and quadratic programming \cite{Zhang2019}, the partial sum of Taylor series is usually applied to approximating a target function so as to achieve an acceptable precision \cite{Mathews2005}. Besides, in recent years, Taylor series have increasingly applied to machine learning and computer vision areas, such as rectifier networks \cite{Balduzzi2017}, saliency map interpretation \cite{Singla2019} and meta learning \cite{Gonzalez2020}.

\section{Motivating Gradient Approximation}
By analysing the cost of PGD adversarial training method, we find that one of the most time-consuming operations is to compute the gradient of adversarial loss [$\nabla_\delta\ell(f_\theta(x+\delta),y)$]. The classifier function $f_\theta$ is a very complex non-linear function (with multi-layer linear combination computation and non-linear activation computation), so it is very costly to compute the gradient of $f_\theta$ for $T$ times in the $T$-step PGD adversarial training method. We find that the adversarial loss can be approximated by partial Taylor series so that we can circumvent the expensive gradient computation and further accelerate adversarial training.

\subsection{Adversarial Loss Approximation}
First of all, we present the third-order Taylor series of a scalar function $f(x)$ with continuous third-order derivatives as follows:
\begin{equation*}
\begin{aligned}
f(x+\triangle x)=&f(x)+\triangle xf^{(1)}(x)+\frac{1}{2}\triangle x^2f^{(2)}(x)\\
&+\frac{1}{6}\triangle x^3f^{(3)}(\xi),
\end{aligned}
\end{equation*}
where $\xi\in(x,x+\triangle x)$ and $\triangle x^3f^{(3)}(\xi)/6$ is the Lagrangian remainder. Because $f^{(3)}(\xi)/6$ is a constant, we can rewrite the remainder as $O(\triangle x^3)=\triangle x^3f^{(3)}(\xi)/6$, which is a function of $\triangle x$. If $\triangle x$ is small enough, we can omit the remainder $O(\triangle x^3)$, so we approximate $f(x+\triangle x)$ as follows:
\begin{equation}\label{eqn.taylor}
f(x+\triangle x)\approx f(x)+\triangle xf^{(1)}(x)+\frac{1}{2}\triangle x^2f^{(2)}(x).
\end{equation}

In our setting, due to the small value of the imperceptible perturbation $\delta$, it is reasonable to approximate the adversarial loss\footnote{For the convenience of representation, we abbreviate adversarial loss $\ell(f_\theta(x+\delta),y)$ as $\ell(x+\delta)$.} $\ell(x+\delta)$ in the matrix form by Formula (\ref{eqn.taylor}):
\begin{equation}\label{eqn.advloss}
\begin{aligned}
\ell(x+\delta)&\approx \ell(x)+\delta^T\nabla_x\ell(x)+\frac{1}{2}\delta^T\nabla_x^2\ell(x)\delta\\
&=\ell(x)+\delta^TJ(x)+\frac{1}{2}\delta^TH(x)\delta,
\end{aligned}
\end{equation}
where $J(x)=\nabla_x\ell(x)$ is the first-order derivative matrix, i.e., the Jacobian matrix; $H(x)=\nabla_x^2\ell(x)$ is the second-order derivative matrix, i.e., the Hessian matrix.

\subsection{Empirical Observations of Adversarial Loss Approximation}
In order to evaluate the effectiveness of the approximation (\ref{eqn.advloss}), we define the residual error between the true loss and the approximate loss as
\begin{equation}\label{eqn.residual}
R(x,\delta) = |\ell(x+\delta)-\ell(x)-\delta^TJ(x)-\frac{1}{2}\delta^TH(x)\delta|,
\end{equation}
where $|\cdot|$ computes the absolute value, and $||\delta||_\infty\le\epsilon$. $H(x)$ is approximated by Equation (\ref{eqn.app_hessian}).

We measure $R(x,\delta)$ of two ResNet models (ResNet-18 and ResNet-50) with RAT on the CIFAR-10 dataset, where the adversary generation is run with 1, 4, 10, 16 steps of PGD attack (\ref{eqn.pgd}). $R(x,\delta)$ is computed throughout training on the training set of CIFAR-10. We used two residual networks with the depth of 18 and 50 as our models. The results are shown in Figure \ref{fig.obser}. At the initial epochs, $R(x,\delta)$ is relatively large, and as the training process goes on, $R(x,\delta)$ decreases. Besides, the more steps the PGD attack has, the larger $R(x,\delta)$ is, while it still converges to very small values, i.e., $7.9\times10^{-4}$ for ResNet-18 and $5.4\times10^{-4}$ for ResNet-50. However, when we train with more PGD steps to generate adversaries, $R(x,\delta)$ is not going to get any larger, e.g., $R(x,\delta)$ of 16-step PGD equals approximately to $R(x,\delta)$ of 10-step PGD.

\section{Adversarial Training with Gradient Approximation }
From the section above, we make the empirical observations that $R(x,\delta)$ decreases as we train with more epochs. In this section, based on the approximation (\ref{eqn.advloss}), we derive an approximation of the gradient of $\ell(x+\delta)$ that can be utilized for accelerating adversarial training.

\subsection{Gradient Approximation of Adversarial Loss}
Instead of computing the partial derivative with respect to $\delta$ of the true adversarial loss, we find the partial derivative of the approximate loss (\ref{eqn.advloss}), so we obtain:
\begin{equation}\label{eqn.approx_grad}
\nabla_\delta\ell(x+\delta)\approx J(x)+H(x)\delta.
\end{equation}

Furthermore, we replace the formula of Line 6 in Algorithm \ref{alg.pgd} with
\begin{equation}\label{eqn.pgd_perturb}
\delta=\delta+\alpha\cdot\mathrm{sign}(J(x)+H(x)\delta).
\end{equation}
In the $T$-step PGD adversarial training, the regular PGD method computes the partial derivative [$\nabla_\delta\ell(x+\delta)$] for $T$ times in a batch. However, except some very simple operations (like addition and multiplication), adversarial training with gradient approximation (\ref{eqn.approx_grad}) only computes the Jacobian matrix and the Hessian matrix of $\ell(x)$ with respect to natural examples $x$ for one time.

\subsection{Hessian Matrix Approximation}
In Equations (\ref{eqn.approx_grad}) and (\ref{eqn.pgd_perturb}), we can easily obtain the Jacobian matrix by backpropagation in adversarial training, while the Hessian matrix is usually difficult to calculate or requires a large amount of calculation. Therefore, we also use approximation method to simplify the calculation.

\textbf{Use $J(x)$ to Approximate $H(x)$.} Referring to the Gauss-Newton method, we use the Jacobian matrix to roughly approximate the Hessian matrix. Gauss-Newton method is a specialized method for minimizing the least squares cost. Given a point $x$, the Gauss-Newton method in our setting is based on the following objective function:
\begin{equation}\label{eqn.gauss_newton1}
\begin{aligned}
&S(\delta)=\frac{1}{2}r^2(\delta), \\
&\mathrm{s.t.}~~r(\delta)=\ell(x+\delta)-\ell(x)\\
&\approx\ell(x)+\delta^T\nabla_x\ell(x)-\ell(x)=\delta^T\nabla_x\ell(x),
\end{aligned}
\end{equation}
where we use the first-order Taylor series to approximate $\ell(x+\delta)$. So the gradient of (\ref{eqn.gauss_newton1}) is $g=r\nabla_x\ell(x)$ (Here, we abbreviate $r(\delta)$ as $r$), and the Hessian matrix is $H(x) = \nabla_x\ell(x)^T\nabla_x\ell(x)+r\nabla_\delta(\nabla_x\ell(x))$. Omitting the second item of $H(x)$, we obtain:
\begin{equation}\label{eqn.app_hessian}
H(x)\approx\nabla_x\ell(x)^T\nabla_x\ell(x).
\end{equation}
Although the objective function (\ref{eqn.gauss_newton1}) is different from the objective function in our task, we can still use (\ref{eqn.app_hessian}) to roughly approximate the Hessian matrix in (\ref{eqn.approx_grad}) and (\ref{eqn.pgd_perturb}) for the purpose of simplifying the calculation.

\textbf{Omit Partial Derivatives to Approximate $H(x)$.} Given an image $x\in\mathbb{R}^{w\times h}$, the Hessian matrix can be written as\footnote{For convenience, we reshape the $w\times h$ image matrix as $wh\times 1$ vector.}
\begin{equation}
H(x)=\left[
\begin{matrix}
\frac{\partial^2\ell}{\partial x_{i}\partial x_{j}}
\end{matrix}
\right]_{wh\times wh},
\end{equation}
where $i$ (or $j$) is the $i$th (or $j$th) element of the image vector and $1\le i,j\le wh$. When $i\ne j$, the partial derivatives $\partial^2\ell/\partial x_{i}\partial x_{j}$ are very difficult to calculate; however, when $i=j$, the second-order derivatives $\partial^2\ell/\partial x_{i}^2$ are easier to obtain. So another idea to approximate $H(x)$ is to omit the partial derivatives ($\partial^2\ell/\partial x_{i}\partial x_{j}$, $i\ne j$) and just set them to zero, and further we obtain the following diagonal matrix for approximating the Hessian matrix:
\begin{equation}\label{eqn.app_hessian2}
H(x)\approx\left[
\begin{matrix}
\frac{\partial^2\ell}{\partial x_{1}^2}&0&\cdots&0\\
0&\frac{\partial^2\ell}{\partial x_{2}^2}&\cdots&0\\
\vdots&\vdots&\ddots&\vdots\\
0&0&\cdots&\frac{\partial^2\ell}{\partial x_{wh}^2}\\
\end{matrix}
\right]_{wh\times wh}.
\end{equation}

Before ending this section, we summarize adversarial training with gradient approximation (GAAT) in Algorithm \ref{alg.gaat} [using (\ref{eqn.app_hessian}) to approximate $H(x)$].

\begin{algorithm}[tb]
\caption{Adversarial training with gradient approximation (GAAT)}
\label{alg.gaat}
\textbf{Input}:  $M$ batches of natural images $(x,y)$\\
\textbf{Parameter}: Training epochs $N$, trained model $f_{\theta}$, learning rate $\gamma$, PGD parameters \{$T$,$\alpha$,$\epsilon$\}\\
\textbf{Output}: Robust model parameters $\theta$\\
\begin{algorithmic}[1]
\FOR{$n=1\dots N$}
\FOR{$i=1\dots M$}
\STATE {$\delta=0$}  {\small\color{gray} \textit{\# or random value}}
\STATE {$J=\nabla_x\ell(x_i)$}  {\small\color{gray} \textit{\# Jacobian matrix}}
\STATE {\small\color{gray} \textit{\# Approximate Hessian matrix}}
\STATE {$H=J^TJ$} {\small\color{gray} \textit{\# or use (10) to approximate $H(x)$}}
\FOR{$t=1\dots T$}
\STATE {$\delta=\delta+\alpha\cdot\mathrm{sign}(J+H\delta)$}
\STATE {$\delta=\max(\min(\delta,\epsilon),-\epsilon)$}  {\small\color{gray} \textit{\# projected into $l_\infty$ ball}}
\ENDFOR
\STATE {\small\color{gray} \textit{\# Update classifier parameters with some optimizer}}
\STATE {$\theta=\theta-\gamma\nabla_\theta\ell(f_\theta(x_i+\delta),y_i)$}
\ENDFOR
\ENDFOR
\RETURN {$\theta$} {\small\color{gray} \textit{\# return parameters of the robust classifier}}
\end{algorithmic}
\end{algorithm}

\begin{table*}[t]
\centering
\begin{tabular}{clcrccc}
\toprule[1pt]
Dataset &Model &Method &Training time &Time saved &Nat acc &Adv acc \\
\midrule
\multirow{8}{*}{CIFAR-10}
& \multirow{2}{*}{VGGNet-16}   &RAT  &421 min &\multirow{2}{*}{46.32\%} &87.00\% &46.64\% \\
&                           &GAAT &226 min  &                         &89.05\% &45.50\% \\
\cline{2-7}
&\multirow{2}{*}{ResNet-18} &RAT  &189 min  &\multirow{2}{*}{30.16\%} &82.54\% &42.10\% \\
&                           &GAAT &132 min  &                         &85.49\% &41.39\% \\
\cline{2-7}
&\multirow{2}{*}{ResNet-50} &RAT  &563 min &\multirow{2}{*}{50.80\%} &86.60\% &46.26\% \\
&                           &GAAT &277 min &                         &87.83\% &44.54\% \\
\cline{2-7}
&\multirow{2}{*}{WideResNet-28x4}&RAT &1345 min &\multirow{2}{*}{60.07\%} &87.72\% &47.07\% \\
&                                &GAAT &537 min & &90.34\% &46.27\% \\
\midrule
\multirow{4}{*}{CIFAR-100}
&\multirow{2}{*}{ResNet-18} &RAT  &189 min &\multirow{2}{*}{29.63\%} &58.24\%  &11.55\% \\
&                           &GAAT &133 min &                         &59.58\%  &9.90\% \\
\cline{2-7}
&\multirow{2}{*}{ResNet-50} &RAT  &564 min &\multirow{2}{*}{51.60\%} &63.86\% &15.30\% \\
&                           &GAAT &273 min &                         &66.66\% &13.25\% \\
\midrule
\multirow{2}{*}{MNIST}
&\multirow{2}{*}{Convolutional ReLU}&RAT  &32 min &\multirow{2}{*}{12.50\%} &99.31\% &96.13\%\\
&                                   &GAAT &28 min &                         &99.30\% &95.13\%\\
\bottomrule[1pt]
\end{tabular}
\caption{Training time and test accuracy (i.e., natural accuracy with natural examples and adversarial accuracy with adversarial examples) for RAT and our method (GAAT) with different models on different datasets.\label{tab.adv}}
\end{table*}

\section{Extensive Experiments and Results}
In this section, we evaluate our proposed method for improving adversarial training efficiency. We compared the training time and test accuracy (i.e., natural accuracy and adversarial accuracy) of the models trained with GAAT as well as the models trained with RAT. To test adversarial accuracy, we used the powerful PGD attack (\ref{eqn.pgd}). MNIST, CIFAR-10 and CIFAR-100 datasets were used in our experiments. Besides, we compared the training time and test accuracy between our method and delayed adversarial training (DAT) \cite{Gupta2020}. Additionally, our method was combined with DAT to further improve the training efficiency of the robust models.

\subsection{Evaluation Setup}
For MNIST, we used the convolutional ReLU model mentioned in \cite{Wong2018}, with two convolutional layers with 16 and 32 $4\times4$ filters each, followed by a fully connected layer with 100 units, with a batch size of 512. For CIFAR-10, we used four popular models, VGGNet-16, ResNet-18, ResNet-50 and WideResNet-28x4 with a batch size of 128. For CIFAR-100, we use ResNet-18 and ResNet-50 models with a batch size of 128.

Generally, we used 200 epochs in total, SGD with momentum 0.9 and weight decay $2\times10^{-4}$ as the optimizer, and an initial learning rate of 0.1, which was decayed by a factor of $0.2$ after epochs 60, 120 and 160. For adversarial training, the parameters are represented in a tuple: \{$T$, $\alpha$, $\epsilon$\}. The adversarial parameters were set to \{10,$1/255$,$8/255$\} for CIFAR-10 and CIFAR-100. For MNIST, the adversarial parameters were set to \{100, 0.01, 0.1\}. All adversarial examples were generated by adding an initial random perturbation $\delta^0$ with $||\delta^0||_\infty\le\epsilon$.

\subsection{Training Time and Test Accuracy}
Table \ref{tab.adv} displays the training time and test accuracy to train all robust models on different datasets. Table \ref{tab.elapsed_time} shows the elapsed time of different models to generate adversarial examples with RAT and GAAT on the CIFAR-10 dataset.

\begin{table}[h!t]
\centering
\begin{tabular}{lrrr}
\toprule[1pt]
RAT (Algorithm \ref{alg.pgd}) & Line 5-8  & Line 6-7 & \\
\midrule
VGGNet-16           & 284 ms     & 27 ms   &          \\
ResNet-18        & 126 ms     & 10 ms   &          \\
ResNet-50        & 389 ms     & 38 ms   &          \\
WideResNet-28x4 & 899 ms    & 67 ms   &         \\
\midrule
\midrule
GAAT (Algorithm \ref{alg.gaat})  & Line 4-10 & Line 8-9 & Line 4-6 \\
\midrule
VGGNet-16           & 140 ms     & 9 ms      & 4 ms      \\
ResNet-18        & 87 ms      & 5 ms      & 5 ms      \\
ResNet-50        & 180 ms     & 12 ms     & 13 ms     \\
WideResNet-28x4 & 327 ms    & 31 ms     & 8 ms      \\
\bottomrule[1pt]
\end{tabular}
\caption{Elapsed time of generating adversarial examples with RAT and GAAT.\label{tab.elapsed_time}}
\end{table}

\begin{figure*}[t]
\centering
\psfrag{Adv acc}[c][c][0.9]{Adv acc/\%}
\psfrag{PGD steps}[c][c][0.9]{PGD steps}
\psfrag{epsilon}[c][c][0.9]{$\epsilon\times 255$}
\psfrag{RAT}[c][c][0.65]{RAT}
\psfrag{DAT}[c][c][0.65]{DAT}
\psfrag{GAAT}[c][c][0.65]{~GAAT}
\psfrag{DAT+GAAT}[c][c][0.65]{DAT+GAAT}
\subfigure{\includegraphics[width=0.81\columnwidth]{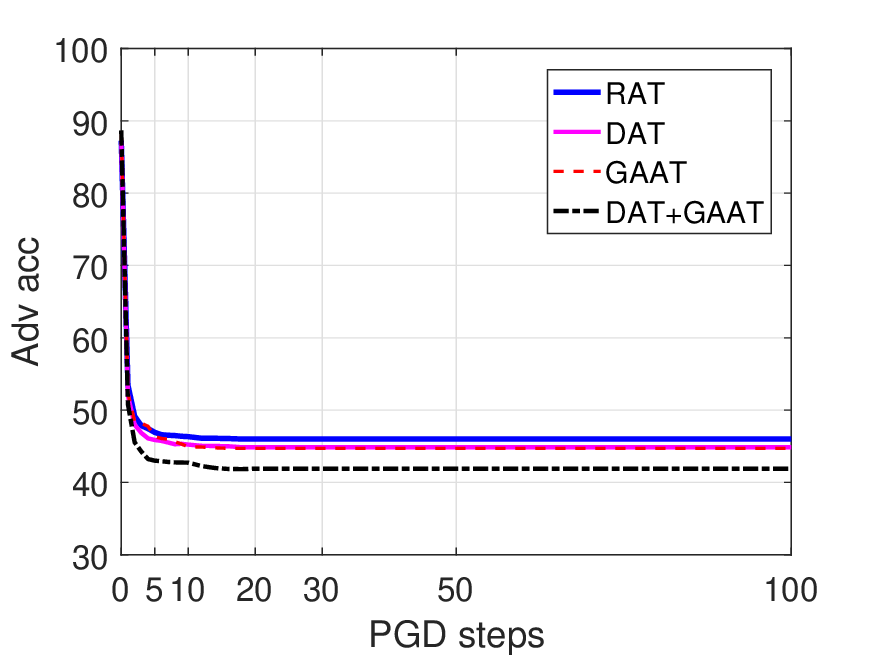}}
\subfigure{\includegraphics[width=0.81\columnwidth]{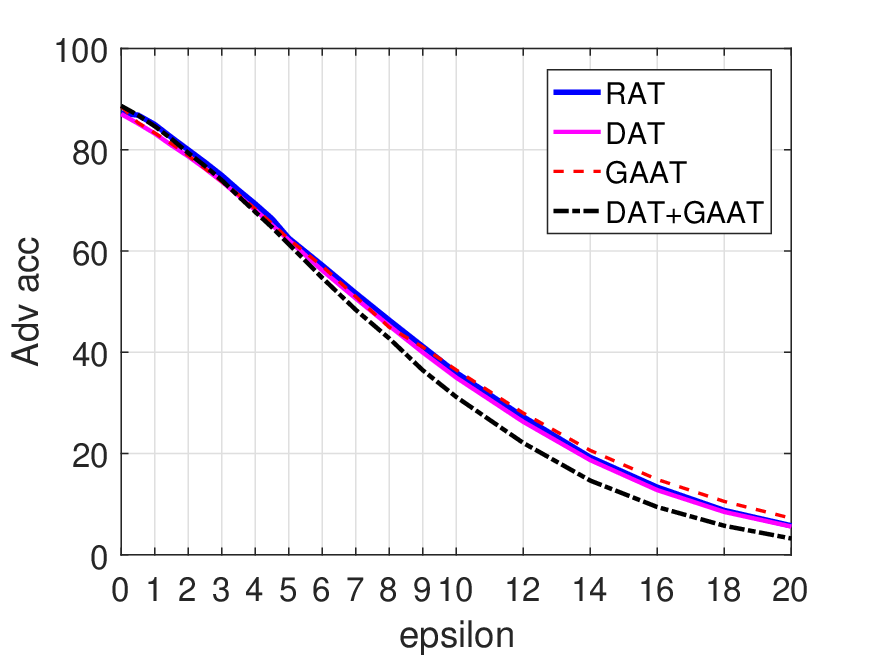}}
\caption{Adversarial accuracy of four well-trained models with the early stopping strategy in Table \ref{tab.combined} on the CIFAR-10 dataset when tested with PGD attacks of different strength.}
\label{fig.attack}
\end{figure*}

\textbf{Training time.} To assess how much training time the approximate gradient can reduce, we conducted experiments to record the elapsed time of generating 10-step adversarial examples with the true gradient and the approximate gradient. The results on the CIFAR-10 dataset with different models are shown in Table \ref{tab.elapsed_time}, where Line 5-8 in Algorithm \ref{alg.pgd} represent generating adversarial examples with RAT; Line 6-7 in Algorithm \ref{alg.pgd} denote one step PGD with RAT; Line 4-10 in Algorithm \ref{alg.gaat} represent generating adversarial examples with GAAT; Line 8-9 in Algorithm \ref{alg.gaat} denote one step PGD with GAAT; and Line 4-6 in Algorithm \ref{alg.gaat} denote computing the Hessian matrix. Evidently, our method greatly reduces the elapsed time of generating adversarial examples, and for larger and more complicated models, e.g., ResNet-50 and WideResNet-28x4, our method reduces the elapsed time by a larger margin. Besides, the time of computing the Hessian matrix (Line 4-6 in Algorithm \ref{alg.gaat}) relates to the depth of models. However, it is negligible compared with the time of generating adversarial examples.

We also evaluated the total training time for RAT and GAAT with different models on different datasets. The results are displayed in Table \ref{tab.adv}, showing that the overhead are remarkably decreased when using our proposed method. For all models, we obtained significant reduction in time overhead. Note that the more complex the model is, the greater overhead reduction our method achieves. For instance, our method saves 60\% of the time overhead when training a robust WideResNet-28x4 model.

\textbf{Test accuracy.} From Table \ref{tab.adv}, we find that the test accuracy of models trained with our method is very close to that of models trained with RAT. In fact, the natural accuracy (tested with natural examples) of models trained with our method is a little higher than that of models trained with RAT, while the adversarial accuracy (tested with adversarial examples) of models trained with our method is a little lower than that of models trained with RAT. The possible reason of these results is that we used an approximate gradient to replace the true gradient. Exactly, the approximate gradient is smaller than the true gradient because the approximate gradient omits the high-order items of the true gradient. Hence, in our method, we used weaker attacks than the attack in RAT to adversarially train a robust model. The little loss of the adversarial accuracy is acceptable, compared with the great time overhead saving.

\subsection{Comparison and Combination with DAT}
Delayed adversarial training (DAT), which uses the natural training to replace the adversarial training for the initial epochs, reduces much time overhead and achieves comparable accuracy. In this subsection, we compare it with our method, and then combine it with our method to further improve adversarial training efficiency. Additionally, early stopping strategy \cite{Rice2020} is regarded as an effective way to avert adversarial overfitting; thus we also used this strategy to save training time and improve model robustness. Table \ref{tab.combined} shows the training time and test accuracy of RAT, DAT, GAAT and DAT+GAAT methods. Here, we used the CIFAR-10 dataset and the ResNet-50 model. In DAT and DAT+GAAT, we switched natural training to adversarial training after epoch 100.

As shown in Table \ref{tab.combined}, GAAT and DAT both significantly reduce the time cost of establishing a robust model, while GAAT can save more time. More intriguingly, when combining DAT with GAAT, DAT+GAAT reduces the time overhead to a greater extent, saving about 70\% of the training time compared with RAT (80\% with early stopping). Besides, by using early stopping strategy, training efficiency and model robustness are both improved.

\begin{table}[t]
\centering
\begin{tabular}{lrcc}
\toprule[1pt]
Method              & Training time & Nat acc   & Adv acc  \\
\midrule
RAT                 & 563 min      & 86.60\%   & 46.26\% \\
~~~+early stopping  & 479 min      & 87.24\%   & 46.33\% \\
\midrule
GAAT                & 277 min      & 87.83\%   & 44.54\%  \\
~~~+early stopping  & 167 min      & 88.53\%   & 45.23\%  \\
\midrule
DAT                 & 307 min      & 86.82\%   & 43.08\%  \\
~~~+early stopping  & 233 min      & 87.04\%   & 44.93\%  \\
\midrule
DAT+GAAT            & 159 min       & 88.48\%   & 42.32\%  \\
~~~+early stopping  & 119 min       & 88.66\%   & 42.73\%  \\
\bottomrule[1pt]
\end{tabular}
\caption{Training time and test accuracy of four adversarial training methods for the ResNet-50 model on the CIFAR-10 dataset.\label{tab.combined}}
\end{table}

\subsection{Generalization to Attacks of Different Strength}
We assessed robustness of the four model in Table \ref{tab.combined} against attacks of different strength which they were not trained to defend against. We varied the number of PGD steps and the value of $\epsilon$ to test attacks. Figure \ref{fig.attack} shows the performances with RAT, DAT, GAAT and DAT+GAAT methods on the CIFAR-10 dataset. The models trained using three efficient methods, i.e., DAT, GAAT and DAT+GAAT, achieve comparable robustness against a wide range of PGD attacks and follows the same pattern as RAT.

\section{Conclusion}
In this paper, we aimed at reducing the computational cost of regular adversarial training. We came up with a new idea about using approximate adversarial gradient to generate adversarial examples. Then, we proposed a variant of the regular adversarial training, adversarial training with gradient approximation, which remarkably improves training efficiency to build up robust deep models. Furthermore, extensive experiments and results showed that our proposed method achieves comparable performances with significantly reducing the computational cost.

%
%
%
%
%
%

\bibliographystyle{named}
\bibliography{ijcairef}

\end{document}